\documentclass{article}
\usepackage{spconf,amsmath,graphicx}
\usepackage{gensymb}
\usepackage{siunitx}
\usepackage{graphicx}
\usepackage{hyperref}

\title{Enhanced Low-resolution LiDAR-Camera Calibration Via Depth Interpolation and Supervised Contrastive Learning}
\name{\textit{Zhikang Zhang}$^{1,\star}$\thanks{$^\star$: equal contribution} \qquad
\textit{Zifan Yu} $^{1,\star}$ \qquad 
\textit{Suya You}$^{2}$ \qquad 
\textit{Raghuveer Rao}$^{2}$ \\
\textit{Sanjeev Agarwal}$^{3}$ \qquad 
\textit{Fengbo Ren}$^{1}$}
\address{$^1$ Arizona State University, Tempe, AZ, \\ \qquad$^2$ U.S. Army DEVCOM Research Laboratory, Adelphi, MD \\ \qquad
$^3$ U.S. Army DEVCOM C5ISR Center, Fort Belvoir, VA}

\begin{document}
%
\maketitle
\begin{abstract}
 Motivated by the increasing application of low-resolution LiDAR recently, we target the problem of low-resolution LiDAR-camera calibration in this work. The main challenges are two-fold: sparsity and noise in point clouds. To address the problem, we propose to apply depth interpolation to increase the point density and supervised contrastive learning to learn noise-resistant features. The experiments on RELLIS-3D demonstrate that our approach achieves an average mean absolute rotation/translation errors of 0.15cm/0.33 \textdegree on 32-channel LiDAR point cloud data, which significantly outperforms all reference methods. 
\end{abstract}
\begin{keywords}
low-resolution point cloud, LiDAR-camera calibration, supervised contrastive learning, image interpolation
\end{keywords}
\section{Introduction}
\begin{figure}[ht]
    \centering
    \includegraphics[width=\linewidth]{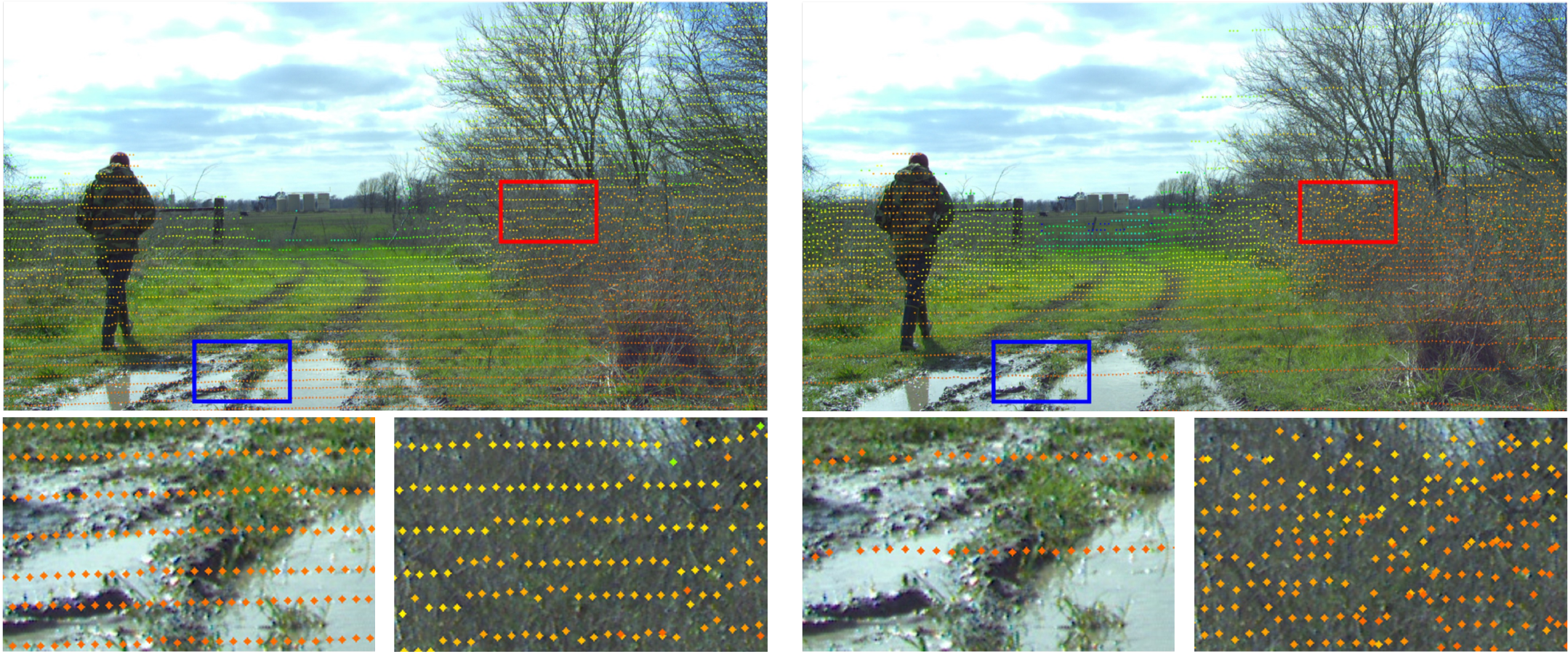}
    \caption{The visual comparison of point clouds from high-resolution LiDAR and low-resolution LiDAR. The point cloud is projected to the image plane and plotted as an overlay layer. The data is from RELLIS-3D\cite{jiang2021rellis}. \textbf{Left}: 64-channel LiDAR. \textbf{Right}: 32-channel LiDAR. \textbf{Blue box}: sparse region. \textbf{Red box}: noisy region.}
    \label{problem}
\end{figure}

Driven by the development of LiDAR technology, the application scenarios of low-resolution LiDAR devices are largely expanded in recent years, such as autonomous driving\cite{lidar1}, geoscience\cite{lidar2}, remote sensing\cite{lidar4}, mobile robotics\cite{lidar6}, etc. To acquire an accurate and informative perception of scanned targets or environments, LiDAR devices are often fused with cameras to utilize rich information of images. The basis of LiDAR-camera fusion is extrinsic calibration, i.e., estimating a relatively rigid body transformation from LiDAR coordinates to camera coordinates, which has been long studied. Conventional calibration methods\cite{tra-calib3,tra-calib4,tra-calib5,tra-calib6,tra-calib7,tra-calib8} are mostly based on explicit targets in a scene, hand-crafted features, or labels of data to build correspondences between point clouds and images, thus are often limited by laborious human interventions and/or applied environments. Recognizing these limitations, recently, deep-learning-based calibration approaches\cite{deepcalib1,deepcalib2,schneider2017regnet,Zhao2021CalibDNNMS,lv2021lccnet} are proposed, which automatically learn features from sensed data and perform calibration in an end-to-end manner. Since the feature learning heavily depends on the quality of data, most works lay the foundation upon highly accurate and noiseless point clouds(Fig.~\ref{problem}) sensed by high-resolution LiDAR, and thus suffer from large performance degradation in low-resolution LiDAR scenarios.
\begin{figure*}[ht]
    \centering
    \includegraphics[width=.90\linewidth]{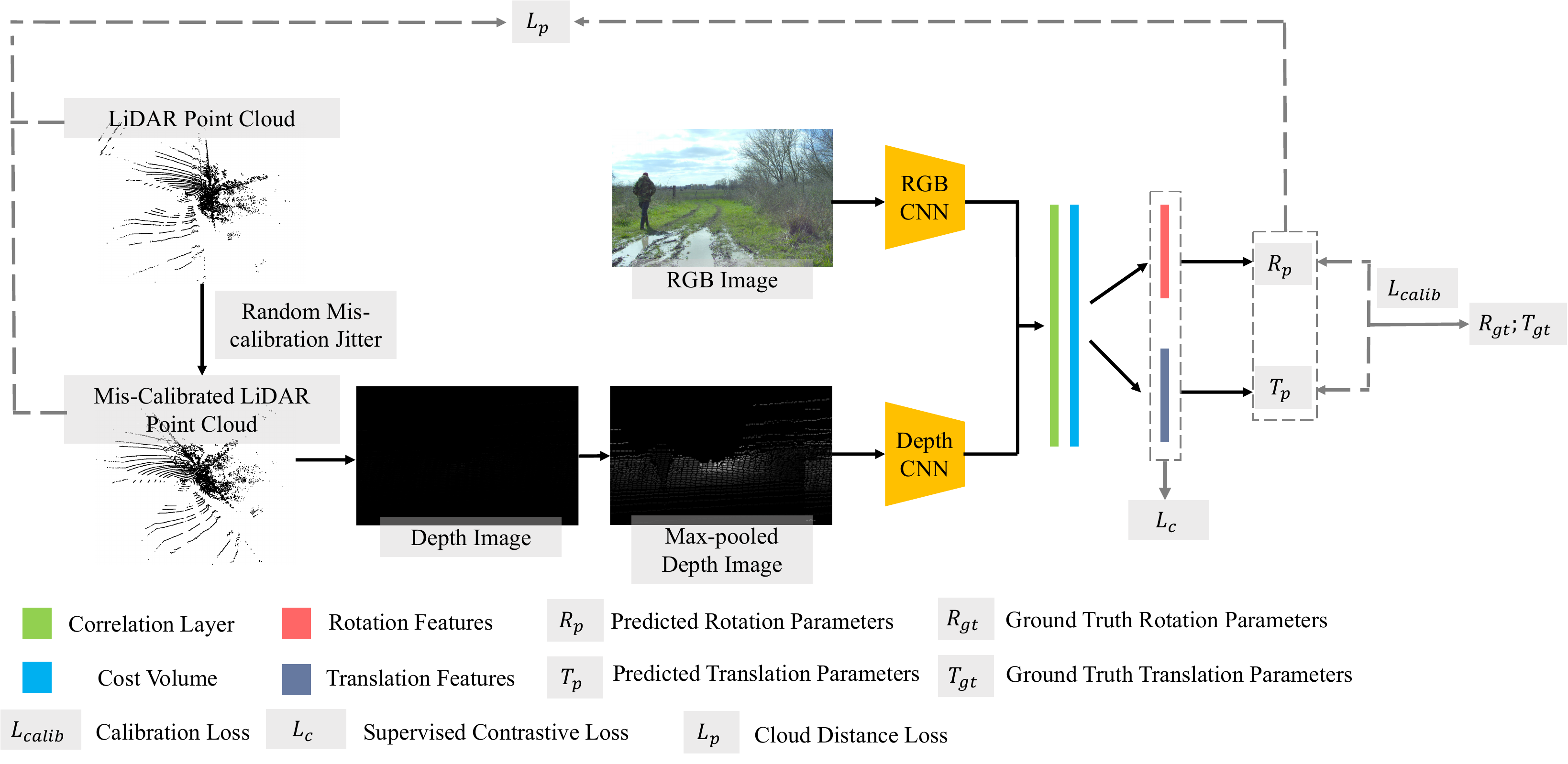}
    \caption{The whole training pipeline of our approach. Given a pair of an \textbf{RGB image} and a \textbf{miscalibrated point cloud}, two \textbf{CNNs} are used to extract features from the RGB image and the \textbf{max-pooled depth image}. Then the extracted features are fed into a \textbf{correlation layer}\cite{Sun2018PWCNetCF} to construct a \textbf{cost volume} for extracting \textbf{rotation features} and \textbf{translation features} which are later used to predict \textbf{rotation parameters} and \textbf{translation parameters}. The rotation parameter is a four-dimensional vector that represents the rotation quaternion. The translation parameter is a three-dimensional translation vector. In the training process, three losses are applied: \textbf{calibration loss} minimizes the distance between predicted calibration parameters and the ground truth. \textbf{Cloud distance loss} minimizes the distance between calibrated point cloud(using predicted calibration parameters) and the ground truth point cloud. \textbf{Supervised contrastive loss} enhances the learned rotation features and translation features to be noise-resistant.}
    \label{pipeline}
\end{figure*}

In this work, we take the state-of-the-art method\cite{lv2021lccnet} as the backbone to apply two effective techniques to enhance the performance of low-resolution LiDAR-camera calibration. We first identify two major challenges to low-resolution LiDAR-camera calibration: sparsity and noise, as shown in Fig.~\ref{problem}. To address the sparsity problem, we apply depth interpolation to increase the density of the point cloud, which inevitably introduces more noise to the point cloud. Then, to address the inherent and introduced noise, we apply supervised contrastive loss on the backbone to learn noise-resistant features for calibration. The extensive experiments on public datasets demonstrate that our approach outperforms all reference methods on low-resolution point clouds by a large margin, which shows strong evidence that our approach is highly effective in addressing the low-resolution LiDAR-camera calibration problem.
Our contributions are summarized as follows:

1. We propose two effective techniques to enhance the deep-learning-based, automatic targetless LiDAR-camera calibration in the low-resolution LiDAR scenario. To the best of our knowledge, this is the first work that targets the low-resolution LiDAR-camera calibration problem. 

2. We demonstrate that supervised contrastive loss can be applied to learn noise-resistant features for LiDAR-camera calibration. 

3. Our approach achieves state-of-the-art performance for low-resolution LiDAR-camera calibration, which sets a strong baseline for this task. 

\section{Methodology}
We use a state-of-the-art method\cite{lv2021lccnet} as the backbone of our approach, as shown in Fig.~\ref{pipeline}. In the inference stage, \cite{lv2021lccnet} runs in two modes: single-stage and multi-stage. In single-stage mode, a single model is trained for a single miscalibration range. In multi-stage mode, multiple models are trained separately for different miscalibration ranges, and the input is calibrated sequentially by models of higher ranges to lower ranges. 

We first identify two main problems for low-resolution LiDAR-camera calibration resulting from point clouds: sparsity and noise. Fig.~\ref{problem} shows a visual comparison of point clouds sensed by a 32-channel LiDAR and a 64-channel LiDAR. Higher sparsity and more noise in low-resolution point clouds lead to more difficulties for calibration since the RGB images and depth images are less correlated, consequently making the constructed cost volume less informative.

\textbf{Depth interpolation.}
To address the sparsity problem, we propose to apply interpolation to depth images before feature extraction. There are a large variety of image interpolation methods, and we choose to use max-pooling, which shows the highest calibration accuracy in Section\ref{di-exp}. Given a depth image with $h$ height and $w$ width and output size of $\hat h$ height and $\hat w$ width, the stride and kernel size are set to $h/\hat{h},w/\hat{w}$ and $h-(\hat h-1)*(h/\hat{h}),w-(\hat w-1)*w/\hat{w}$, respectively, following the widely used adaptive max-pooling design\cite{maxpool}.

\textbf{Supervised contrastive learning.} 
Depth interpolation inevitably introduces more noise to the point cloud since a large amount of fake 3d points are added. To learn noise-resistant features, we hypothesize that learned features should satisfy three conditions: 1. Rotation features(Fig.~\ref{pipeline}, red block) only retain information related to rotation parameters. 2. Translation features(Fig.~\ref{pipeline}, purple block) only retain information related to translation parameters. 3. Both rotation and translation features do not retain data-dependent(either image or point cloud)information to avoid over-fitting.

Following the three conditions, we propose to add supervised contrastive loss(SCL)\cite{Khosla2020SupervisedCL} in addition to calibration loss and cloud distance loss(both defined in original paper\cite{lv2021lccnet}). SCL is defined as
\begin{equation}
{L}^{s u p}=\sum_{i\in I}\frac{-1}{|P(i)|}\sum_{p\in P(i)}\log\frac{\exp\left(z_{i}\cdot z_{p}/\tau\right)}{\exp\left(z_{i}\cdot z_{a}/\tau\right)}
  \label{supervised1}
\end{equation}
where $P(i)\equiv\{p\in A(i):\tilde{y}_{p}=\tilde{y}_{i}\}$ is the set of indices of all positives samples distinct from $i$ within the mini-batch, $|P(i)|$ is its cardinality, and $z_i$ is the feature of the corresponding sample, as detailed in \cite{Khosla2020SupervisedCL}. Despite its complicated mathematical form, SCL can be implemented as a function that takes in a batch of features and the same number of numerical labels while outputting a singular loss value, i.e., 
\begin{equation}
    loss=SCL([f_1,\cdots,f_b],[l_1,\cdots,l_b])
\end{equation} where $f_i$ is the feature, $l_i$ is the corresponding label, and $b$ is the batch size, as implemented in \cite{subcontrast}. Then in the training process, features with the same labels are pulled together while simultaneously, features with different labels are pushed apart. 

To adapt SCL for enhanced feature learning, we generate features and labels in the following strategy for each batch: with a batch size of $b$, given a batch of training samples containing 4-tuples $(I_k, P_k, R_k, T_k)$ of RGB image, point cloud, random rotation, and random translation, $1 \le k \le b$, we first compose a new batch as inputs consisting of all possible 4-tuple combinations of $I_k, P_k, R_k, T_k$ while always keeping $I_k$ and $P_k$ paired, as shown in Table~\ref{permutation}. As such, the new batch size is $b^3$. Then we assign two groups of labels to generated rotation features $R^f_k$ and translation features $T^f_k$, respectively. For rotation features, the same labels are assigned if and only if they have the same rotation parameters. For translation features, their labels are assigned in a similar manner to correspond to translation parameters. Then, two SCL functions are used to take in rotation and translation features and the corresponding labels, respectively. Through this process, three conditions can be satisfied in the following sense: 1. Rotation features $R^f$ are pushed closer if and only if their rotation parameters $R$ are the same. 2. Translation features $T^f$ are pushed closer if and only if their translation parameters $T$ are the same. 3. Rotation features and translation features are less affected by solely changing the input images and point cloud pairs without changing the calibration parameters. 
\begin{table}[ht]
\centering
\begin{tabular}{|cccc|cc|cc|}
\hline
\multicolumn{4}{|c|}{Composed inputs}                                                                               & \multicolumn{2}{c|}{Features}               & \multicolumn{2}{c|}{Labels}                 \\ \hline
\multicolumn{1}{|c|}{Image} & \multicolumn{1}{c|}{PC} & \multicolumn{1}{c|}{RO} & TR & \multicolumn{1}{c|}{RO} & TR & \multicolumn{1}{c|}{RO} & TR \\ \hline
\multicolumn{1}{|c|}{$I_1$}    & \multicolumn{1}{c|}{$P_1$}          & \multicolumn{1}{c|}{$R_1$}       & $T_1$       & \multicolumn{1}{c|}{$R^f_1$}      & $T^f_1$         & \multicolumn{1}{c|}{1}        & 1           \\ \hline
\multicolumn{1}{|c|}{$I_1$}    & \multicolumn{1}{c|}{$P_1$}          & \multicolumn{1}{c|}{$R_1$}       & $T_2$       & \multicolumn{1}{c|}{$R^f_2$}      & $T^f_2$         & \multicolumn{1}{c|}{1}        & 2           \\ \hline
\multicolumn{1}{|c|}{...}   & \multicolumn{1}{c|}{...}         & \multicolumn{1}{c|}{...}      & ...      & \multicolumn{1}{c|}{...}      & ...         & \multicolumn{1}{c|}{...}         & ...            \\ \hline
\multicolumn{1}{|c|}{$I_1$}    & \multicolumn{1}{c|}{$P_1$}          & \multicolumn{1}{c|}{$R_2$}       & $T_1$       & \multicolumn{1}{c|}{$R^f_{b+1}$}     & $T^f_{b+1}$        & \multicolumn{1}{c|}{2}        & 1           \\ \hline
\multicolumn{1}{|c|}{...}   & \multicolumn{1}{c|}{...}         & \multicolumn{1}{c|}{...}      & ...      & \multicolumn{1}{c|}{...}         &  ...           & \multicolumn{1}{c|}{...}         &   ...          \\ \hline
\multicolumn{1}{|c|}{$I_2$}    & \multicolumn{1}{c|}{$P_2$}          & \multicolumn{1}{c|}{$R_1$}       & $T_1$       & \multicolumn{1}{c|}{$R^f_{b*b+1}$}    & $T^f_{b*b+1}$     & \multicolumn{1}{c|}{1}        & 1           \\ \hline
\multicolumn{1}{|c|}{...}   & \multicolumn{1}{c|}{...}         & \multicolumn{1}{c|}{...}      & ...      & \multicolumn{1}{c|}{...}      & ...         & \multicolumn{1}{c|}{...}      & ...         \\ \hline
\multicolumn{1}{|c|}{$I_b$}    & \multicolumn{1}{c|}{$P_b$}          & \multicolumn{1}{c|}{$R_b$}       & $T_b$       & \multicolumn{1}{c|}{$R^f_{b*b*b}$}    & $T^f_{b*b*b}$       & \multicolumn{1}{c|}{b}        & b           \\ \hline
\end{tabular}%
\caption{The composed input batch and assigned labels for supervised contrastive learning. Original batch size: $b$. PC: point cloud. RO: rotation. TR: translation. $I_k$: $k^{th}$ image in the batch. $P_k$: $k^{th}$ point cloud. $R_k$: $k^{th}$ random rotation parameters. $T_k$: $k^{th}$ random translation parameters. $R^f_k$: $k^{th}$ generated rotation feature. $T^f_k$: $k^{th}$ generated translation feature.}
\label{permutation}
\end{table}

\section{Experiments}
We use RELLIS-3D\cite{jiang2021rellis} dataset for evaluation. RELLIS-3D contains point clouds sensed by a 32-channel LiDAR and 64-channel LiDAR in off-road environments. 32-channel point clouds are treated as low-resolution data. The split of the dataset follows the official split in \cite{jiang2021rellis}, with 7800 training samples, 2413 validation samples, and 3343 testing samples. The training of LCCNet is following the setups in the original paper\cite{lv2021lccnet}, and the miscalibration ranges are set to $150cm/20\degree$, $100cm/ 10\degree$, $50cm/5\degree$, $20cm/2\degree$, and $10cm/1\degree$, which is consistent with \cite{lv2021lccnet} and \cite{schneider2017regnet}. The evaluation metrics are mean absolute translation error $(x,y,z)$, mean absolute rotation errors $(roll, pitch, yaw)$, averaged translation error $(x+y+z)/3$ and averaged rotation error $(roll+pitch+yaw)/3$.

\textbf{Quantify calibration performance degradation.}
We first train two multi-stage LCCNet\cite{lv2021lccnet} on point clouds of 32 channels and 64 channels, respectively. The experiment results are shown in Table~\ref{degrad}. The average translation error and rotation error increase two to four times on 32-channel data compared with the same model trained on 64-channel data.
\begin{table}[htb]
\centering
\begin{tabular}{|c|c|c|c|c|c|c|}
\hline
Channel & X    & Y    & Z    & Roll & Pitch & Yaw  \\ \hline
64      & 0.66 & 0.71 & 0.25 & 0.12 & 0.14  & 0.09 \\ \hline
32      & 2.6  & 2.6  & 2.78 & 0.22 & 0.16  & 0.27 \\ \hline
\end{tabular}
\caption{Quantified performance degradation on low-resolution(32-channel) LiDAR. Unit: cm or \textdegree}
\label{degrad}
\end{table}

\textbf{Depth interpolation.}
\label{di-exp}
We compare max-pooling against three candidate image interpolation methods: average-pooling, linear interpolation, and nearest neighbor interpolation, as well as the original LCCNet approach. The single-stage \cite{lv2021lccnet} is trained at the miscalibration range of $150cm/20\degree$. The experiment results are shown in Table~\ref{interpolation-exp}. The model trained with max-pooling achieves the lowest calibration errors among all interpolation methods. We choose to employ max-pooling to interpolate depth images in the following experiments. Be noted the rotation errors of all four interpolation methods are slightly higher than the original model, which can be attributed to the fake points added to the depth image through interpolation.
\begin{table}[htb]
\centering
\begin{tabular}{|c|c|c|c|c|c|}
\hline
   & original & linear & \begin{tabular}[c]{@{}c@{}}average\\ pooling\end{tabular} & \begin{tabular}[c]{@{}c@{}}max\\ pooling\end{tabular} & \begin{tabular}[c]{@{}c@{}}nearest \\ neighbour\end{tabular} \\ \hline
TR & 54.13    & 71.78  & 43.00                                                     & 40.84                                                 & 57.99                                                        \\ \hline
RO & 1.02     & 3.95   & 4.00                                                      & 3.34                                                  & 4.23                                                         \\ \hline
\end{tabular}
\caption{Comparison against various image interpolation methods. TR: averaged translation error(unit: cm). RO: averaged rotation error (unit: \textdegree).}
\label{interpolation-exp}
\end{table}

\textbf{Supervised contrastive learning.}
We validate the effectiveness of SCL by training single-stage model on all five different miscalibration ranges. As experiment results in Table~\ref{single} show, with max-pooling applied, the averaged translation and rotation errors at most ranges are significantly reduced compared to the original approach. In addition, with SCL being applied, the calibration error is further reduced by 3.95cm/0.25\textdegree on average. The experiment results validate our hypothesis that SCL can enhance feature learning of calibration.

\begin{table}[htb]
\centering
\begin{tabular}{|c|c|c|c|}
\hline
range  & original   & MP & MPSCL        \\ \hline
150/20 & 54.12/1.02 & 40.84/3.34  & 26.86/2.61 \\ \hline
100/10 & 17.13/0.64 & 11.82/0.91  & 9.52/0.61  \\ \hline
50/5   & 8.78/0.50  & 5.08/0.36   & 3.17/0.25  \\ \hline
20/2  & 4.05/0.41  & 3.17/0.23   & 1.75/0.18  \\ \hline
10/1   & 2.11/0.21  & 0.98/0.17   & 0.84/0.12  \\ \hline
\end{tabular}
\caption{Calibration performance comparison at different miscalibration ranges. Original: the original model. MP: with max-pooling applied. MPSCL: with both max-pooling and SCL applied. Unit: cm/\textdegree}
\label{single}
\end{table}

\textbf{Comparison against reference methods.}
We further evaluate the performance of our approach(multi-stage model trained at ranges of $150cm/20\degree$, $100cm/ 10\degree$, $50cm/5\degree$, $20cm/2\degree$, and $10cm/1\degree$ with max-pooling and SCL applied, denoted as MPSCL) by comparing it against multiple reference methods. The miscalibration range for evaluation is set to $150cm/ 20\degree$. The reference methods are original multi-stage LCCNet, Regnet\cite{schneider2017regnet}, and CalibDNN\cite{Zhao2021CalibDNNMS}(For CalibDNN, the range is set to $20cm/10\degree$ to be consistent with original work). The experiment results are shown in Table~\ref{mainresult}. MPSCL achieves the highest performance on all evaluation metrics. Compared with LCCNet, the averaged translation error and rotation error are reduced by 87\%(2.66cm to 0.33cm) and 28\%(0.21\textdegree to 0.15\textdegree), respectively. Compared with the two reference methods, the calibration errors of MPSCL are at least one order of magnitude lower, which is strong evidence that MPSCL can effectively perform LiDAR-camera calibration in low-resolution LiDAR scenarios. 
\begin{table}[htb]
\centering
\resizebox{\linewidth}{!}{%
\begin{tabular}{|c|c|c|c|c|}
\hline
              & RegNet & CalibDNN & LCCNet & MPSCL \\ \hline
X     & 58.70  & 10.43    & 2.60   & 0.23         \\ \hline
Y     & 32.30  & 14.59    & 2.60   & 0.45         \\ \hline
Z     & 50.73  & 10.77    & 2.78   & 0.30         \\ \hline
Average  & 47.24  & 11.93    & 2.66   & 0.33         \\ \hline
Roll  & 4.00   & 1.11     & 0.22   & 0.14         \\ \hline
Pitch & 8.23   & 4.15     & 0.16   & 0.13         \\ \hline
Yaw   & 5.42   & 2.05     & 0.27   & 0.17         \\ \hline
Average  & 5.88   & 2.44     & 0.21   & 0.15         \\ \hline
\end{tabular}
}
\caption{The performance evaluation of MPSCL against reference methods. Unit: cm or \textdegree.}
\label{mainresult}
\end{table}

\textbf{Performance on subsampled point clouds.}
To our knowledge, there is no public dataset for the LiDAR-camera calibration problem with a resolution below 32 channels. To evaluate the performance of our approach in extreme cases, we perform subsampling on point clouds to simulate lower-resolution LiDAR scenarios. We test with three subsampling rates: 2, 4, and 8. The point cloud is uniformly subsampled. The miscalibration range is set to $150cm/20\degree$. The experiment results are shown in Table~\ref{subsample}. MPSCL again shows significantly higher performance than all reference methods. Even at a subsampling rate of 8, the averaged translation/rotation errors are only 4.28cm/1.24\textdegree. Compared with LCCNet, the average reduction in average translation/rotation errors is 19.43cm/0.03\textdegree. Compared with RegNet and CalibDNN, the average reduction is 51.09cm/5.25\textdegree and 7.34cm/0.91\textdegree, respectively. This is further evidence that MPSCL can well address the LiDAR-camera calibration problem in low-resolution LiDAR scenarios. 

\begin{table}[htb]
\centering
\begin{tabular}{|c|c|c|c|}
\hline
Subsample rate   & 2      & 4      & 8      \\ \hline
RegNet       & 49.71/5.96 & 54.19/6.21 & 57.80/6.76 \\ \hline
CalibDNN     & 11.45/1.55 & 10.31/1.73 & 8.40/2.65  \\ \hline
LCCNet       & 4.23/0.57  & 22.73/0.51 & 39.45/2.20 \\ \hline
MPSCL & 1.21/0.04  & 2.65/1.91  & 4.28/1.24  \\ \hline
\end{tabular}
\caption{Performance evaluation on subsampled point clouds. Unit: cm/\textdegree}
\label{subsample}
\end{table}
\section{Conclusion}
We propose an effective approach for low-resolutioin LiDAR-camera calibration. We first identify two main challenges in this problem resulting from low-resolution data: sparsity and noise. Then, we take \cite{lv2021lccnet} as the backbone to apply max pooling to interpolate depth images and supervised contrastive loss to tackle noises, which eventually leads to a highly effective approach for low-resolution LiDAR-camera calibration. The extensive experiments on RELLIS-3D against reference methods demonstrate that our approach can achieve superior performance in calibration, even for extreme cases.
\bibliographystyle{IEEEbib}
\bibliography{refs}

\begin{thebibliography}{10}

\bibitem{lidar1}
Ying Li, Lingfei Ma, Zilong Zhong, Fei Liu, Dongpu Cao, Jonathan Li, and
  Michael~A. Chapman,
\newblock ``Deep learning for lidar point clouds in autonomous driving: A
  review,''
\newblock {\em IEEE Transactions on Neural Networks and Learning Systems}, vol.
  32, pp. 3412--3432, 2021.

\bibitem{lidar2}
Gregor Luetzenburg, Aart Kroon, and Anders~Anker Bj{\o}rk,
\newblock ``Evaluation of the apple iphone 12 pro lidar for an application in
  geosciences,''
\newblock {\em Scientific Reports}, vol. 11, 2021.

\bibitem{lidar4}
Christoph Gollob, Tim Ritter, Ralf Kra{\ss}nitzer, Andreas Tockner, and Arne
  Nothdurft,
\newblock ``Measurement of forest inventory parameters with apple ipad pro and
  integrated lidar technology,''
\newblock {\em Remote. Sens.}, vol. 13, pp. 3129, 2021.

\bibitem{lidar6}
Tao Yang, You Li, Cheng Zhao, Dexin Yao, Guanyin Chen, Li~Sun, Tomas Krajnik,
  and Zhi Yan,
\newblock ``3d tof lidar in mobile robotics: A review,''
\newblock {\em arXiv preprint arXiv:2202.11025}, 2022.

\bibitem{tra-calib3}
Tekla T{\'o}th, Zolt{\'a}n Pusztai, and Levente Hajder,
\newblock ``Automatic lidar-camera calibration of extrinsic parameters using a
  spherical target,''
\newblock in {\em 2020 IEEE International Conference on Robotics and Automation
  (ICRA)}. IEEE, 2020, pp. 8580--8586.

\bibitem{tra-calib4}
Felix Igelbrink, Thomas Wiemann, Sebastian P{\"u}tz, and Joachim Hertzberg,
\newblock ``Markerless ad-hoc calibration of a hyperspectral camera and a 3d
  laser scanner,''
\newblock in {\em International Conference on Intelligent Autonomous Systems}.
  Springer, 2018, pp. 748--759.

\bibitem{tra-calib5}
Peng Jiang, Philip Osteen, and Srikanth Saripalli,
\newblock ``Semcal: Semantic lidar-camera calibration using neural mutual
  information estimator,''
\newblock in {\em 2021 IEEE International Conference on Multisensor Fusion and
  Integration for Intelligent Systems (MFI)}. IEEE, 2021, pp. 1--7.

\bibitem{tra-calib6}
Xinyu Zhang, Shifan Zhu, Shichun Guo, Jun Li, and Huaping Liu,
\newblock ``Line-based automatic extrinsic calibration of lidar and camera,''
\newblock in {\em 2021 IEEE International Conference on Robotics and Automation
  (ICRA)}. IEEE, 2021, pp. 9347--9353.

\bibitem{tra-calib7}
Huai Yu, Weikun Zhen, Wen Yang, and Sebastian Scherer,
\newblock ``Line-based 2-d--3-d registration and camera localization in
  structured environments,''
\newblock {\em IEEE Transactions on Instrumentation and Measurement}, vol. 69,
  no. 11, pp. 8962--8972, 2020.

\bibitem{tra-calib8}
J~Per{\v{s}}i{\'c}, L~Petrovi{\'c}, I~Markovi{\'c}, and I~Petrovi{\'c},
\newblock ``Online multi-sensor calibration based on moving object tracking,''
\newblock {\em Advanced Robotics}, vol. 35, no. 3-4, pp. 130--140, 2021.

\bibitem{deepcalib1}
Ganesh Iyer, R~Karnik Ram, J~Krishna Murthy, and K~Madhava Krishna,
\newblock ``Calibnet: Geometrically supervised extrinsic calibration using 3d
  spatial transformer networks,''
\newblock in {\em 2018 IEEE/RSJ International Conference on Intelligent Robots
  and Systems (IROS)}. IEEE, 2018, pp. 1110--1117.

\bibitem{deepcalib2}
Kaiwen Yuan, Zhenyu Guo, and Z~Jane Wang,
\newblock ``Rggnet: Tolerance aware lidar-camera online calibration with
  geometric deep learning and generative model,''
\newblock {\em IEEE Robotics and Automation Letters}, vol. 5, no. 4, pp.
  6956--6963, 2020.

\bibitem{schneider2017regnet}
Nick Schneider, Florian Piewak, Christoph Stiller, and Uwe Franke,
\newblock ``Regnet: Multimodal sensor registration using deep neural
  networks,''
\newblock in {\em 2017 IEEE intelligent vehicles symposium (IV)}. IEEE, 2017,
  pp. 1803--1810.

\bibitem{Zhao2021CalibDNNMS}
Ganning Zhao, Jiesi Hu, Suya You, and C.~C.~Jay Kuo,
\newblock ``Calibdnn: multimodal sensor calibration for perception using deep
  neural networks,''
\newblock in {\em Defense + Commercial Sensing}, 2021.

\bibitem{lv2021lccnet}
Xudong Lv, Boya Wang, Ziwen Dou, Dong Ye, and Shuo Wang,
\newblock ``Lccnet: Lidar and camera self-calibration using cost volume
  network,''
\newblock in {\em Proceedings of the IEEE/CVF Conference on Computer Vision and
  Pattern Recognition}, 2021, pp. 2894--2901.

\bibitem{Sun2018PWCNetCF}
Deqing Sun, Xiaodong Yang, Ming-Yu Liu, and Jan Kautz,
\newblock ``Pwc-net: Cnns for optical flow using pyramid, warping, and cost
  volume,''
\newblock {\em 2018 IEEE/CVF Conference on Computer Vision and Pattern
  Recognition}, pp. 8934--8943, 2018.

\bibitem{jiang2021rellis}
Peng Jiang, Philip Osteen, Maggie Wigness, and Srikanth Saripalli,
\newblock ``Rellis-3d dataset: Data, benchmarks and analysis,''
\newblock in {\em 2021 IEEE international conference on robotics and automation
  (ICRA)}. IEEE, 2021, pp. 1110--1116.

\bibitem{maxpool}
``Adaptive max pooling,''
  \url{https://pytorch.org/docs/stable/generated/torch.nn.AdaptiveMaxPool2d.html},
\newblock Accessed: 2022-10-19.

\bibitem{Khosla2020SupervisedCL}
Prannay Khosla, Piotr Teterwak, Chen Wang, Aaron Sarna, Yonglong Tian, Phillip
  Isola, Aaron Maschinot, Ce~Liu, and Dilip Krishnan,
\newblock ``Supervised contrastive learning,''
\newblock {\em ArXiv}, vol. abs/2004.11362, 2020.

\bibitem{subcontrast}
``Supcontrast,'' \url{https://github.com/HobbitLong/SupContrast},
\newblock Accessed: 2022-10-19.

\end{thebibliography}
\end{document}